\documentclass[10pt,twoside]{article}
\usepackage{Latex-document}
\almvol{00}
\almttone{Frontiers of Science Awards 2024}
\almtttwo{for Math/CS/Phys}
\firstpage{1}
\usepackage[english]{babel}
\usepackage[utf8]{inputenc}

\usepackage{multicol}
\usepackage[a4paper, total={6in, 8in}]{geometry}

\usepackage{tabu}
\usepackage{amsmath}
\usepackage{amssymb}
\usepackage{graphicx}
\usepackage{wrapfig}
\usepackage{bm}

\numberwithin{equation}{section}

\begin{document}

\markboth{\hfill{\rm Raissi, Perdikaris, Ahmadi and  Karniadakis} \hfill}{\hfill {\rm Physics-Informed Neural Networks and Extensions \hfill}}

\title{Physics-Informed Neural Networks and Extensions}

\author{Maziar Raissi$^{1}$, Paris Perdikaris$^{2}$, Nazanin Ahmadi$^{3}$,  and George Em Karniadakis$^{4}$}

\begin{abstract}
In this paper, we review the new method Physics-Informed Neural Networks (PINNs) that has become the main pillar in scientific machine learning, we present recent practical extensions, and provide a specific example in data-driven discovery of governing differential equations.
\end{abstract}

\maketitle

\setcounter{tocdepth}{1}
\tableofcontents

\section{Overview}
\subsection{Physics-Informed Neural Networks - PINNs} 

In the last 50 years there has been a tremendous success in solving numerically PDEs using finite elements, spectral, and even meshless methods.
Yet, in many 
real cases, we still cannot incorporate seamlessly (multi-fidelity) data into existing algorithms, and for industrial-complexity applications the mesh generation is time consuming and still an art. Moreover, solving {\em inverse problems}, e.g., for material properties, is often prohibitively expensive and requires different formulations and new computer codes. In recent years, uncertainty
quantification (UQ) of simulations has led to highly parametrized formulations that may include 100s of uncertain parameters for complex problems rendering such computations infeasible in practice. Finally, existing computer programs for engineering  applications have more than 100,000 lines of code, making it almost impossible to maintain and update them from one generation to the next. 
To this end, physics-informed learning, i.e., integrating seamlessly data and mathematical models, and implementing them using physics-informed neural networks (PINNs) \cite{raissi2017physics_I}, \cite{raissi2017physics_II}, \cite{raissiJCP2019}
is a paradigm shift in defining the main thrust in scientific machine learning (SciML).

\begin{wrapfigure}{r}{6.0 cm}
\includegraphics[width=\linewidth]{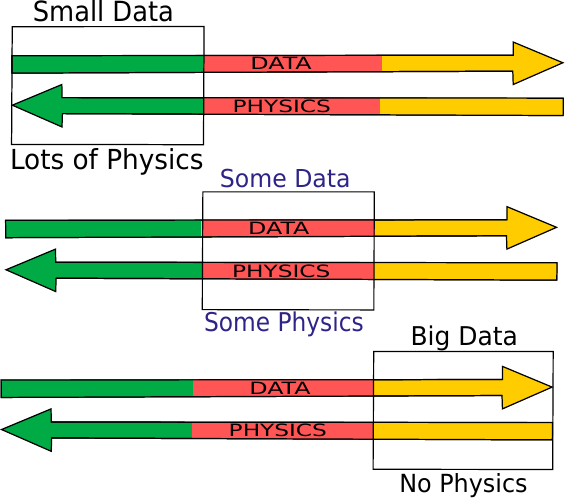}
% \vspace{-0.2in}
\caption{
\small{Schematic to illustrate three possible categories of physical problems and associated available data: 
Physics-informed neural networks can integrate seamlessly data and parameterized PDEs, including models with missing physics, in a unified way expressed compactly using automatic differentiation and PDE-induced neural networks.}}
\label{fig:data} 
\vspace*{-0.2in}
\end{wrapfigure}

\begin{wrapfigure}{r}{0.5\textwidth}
\includegraphics[trim={4cm 3cm 2cm 3cm}, clip, width=0.65\textwidth]{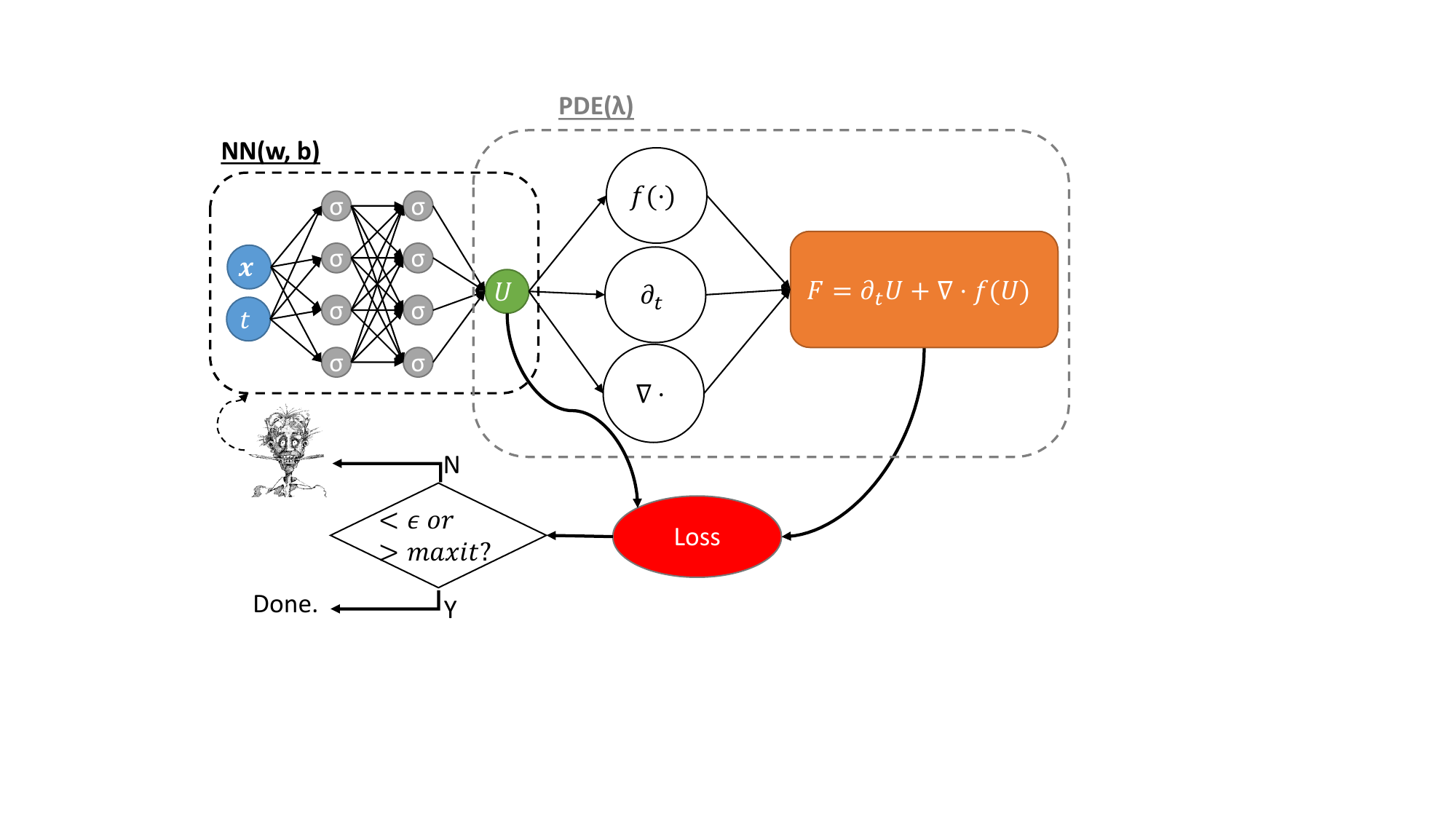} 
\vspace*{-0.4in}
\caption{\small Basic structure of PINN for conservation laws. The left (physics uninformed) network represents the PDE solution ``U(x,t)" while the right (physics informed) network describes the PDE residual ``F(x,t)". Currently, optimization is done by a human-in-the-loop empirically based on trial and error. Note that the ``U" architecture is a fully-connected DNN here (or CNN, RNN, hybrid), while the ``F" architecture is dictated by the PDE and is, in general, not possible to visualize explicitly. Its depth is proportional to the highest derivative order in the PDE times the depth of the uninformed ``U" DNN.}
\label{fig:Af2}
\vspace*{-0.2in}
\end{wrapfigure}

The specific data-driven approach to modeling physical systems depends crucially on the amount of data available as well as on the complexity of the system itself, as illustrated in Fig.~\ref{fig:data}. The classical paradigm 
%for which many different numerical methods have been developed over the last 50 years 
is shown on the top of Fig. \ref{fig:data}, where we assume that the only data available are the boundary conditions (BC) and initial conditions (IC) while the specific governing partial differential equations (PDEs) and related parameters are precisely known. On the other extreme (lower plot), we may have a lot of data, e.g. in the form of time series, but we may not know the governing physical law, e.g. the underlying PDE, at the continuum level.
For most real applications, the most interesting category is sketched in the middle plot, where we assume that we  know the physics partially, e.g., we know the conservation law but not the constitutive relationship, but we have several scattered measurements (of a primary or auxiliary state) that we can use to infer parameters and even  missing {\em functional terms} in the PDE while simultaneously recover the solution. 
%%%%%%%%%%%%%%%%%%%%%%%%%%%%%%%%%%%%%%%%%%%%%%%%%%
%%%%%%%%%%%%%%%%%%%%%%%%%%%%%%%%%%%%%%%%%%%%%%%%%%
%It is clear that this middle category is the most general case, and in fact it is representative of the other two categories, if the measurements are too few or too many. 
This is the mixed case, which may lead to significantly more complex scenarios, where the solution is a stochastic process due to stochastic excitation or an uncertain material property. Hence, we can employ stochastic PDEs (SPDEs) to represent these stochastic solutions and uncertainties. The concept of PINNs was introduced in~\cite{raissi2017physics_I}\\ \cite{raissiJCP2019}, see Fig.~\ref{fig:Af2}, and is being adopted across many scientific domains. 
%Finally, there are many problems involving long-range spatio-temporal interactions, e.g. turbulence, visco-elasto-plastic materials or other anomalous transport processes, where nonlocal/fractional calculus and fractional PDEs (FPDEs) may be the proper mathematical model to adequately describe such phenomena~\cite{Mainardi_2010} as they exhibit a rich expressivity not unlike that of deep neural networks (DNNs).

%{\bf Learning and Meta-Learning.} 
%Decision making and planning in many military and civilian applications relies on “dinky, dynamic,
%dirty, deceptive data” (5D). Although machine learning techniques have been proposed to work
%on pristine and copious amounts of data, 5D and scientific applications require rethinking current
%approaches. To this end, 

%This is a radical departure from traditional machine learning techniques as applied
%to data today – the kind of paradigm shift needed in a 5D world and scientific applications. 
In particular, prior physics-based information ({\em even imperfect}) -- in the form of conservation laws,
dynamic and kinematic constraints -- regularizes a deep neural network (DNN) so it can learn from ``small" and noisy data. In the early works, we 
applied PINNs to a number of nonlinear problems in physics and mechanics, and have demonstrated that PINNs converge to accurate solutions of
PDEs by leveraging or discovering the ``hidden physics" of the data without using any regular grids or discretization in space-time either for forward or for inverse problems. (The use of automatic differentiation employed in the DNN backpropagation is also used to implement the differential operators in the PDEs.) Moreover, PINNs seem to tackle ill-posed (in the classical sense) forward and inverse problems, where no IC/BC are specified or some of the parameters in the PDEs are unknown -- scenarios where classical numerical methods may fail.

\subsection{Extensions of PINNs}

%- weights: Paris, TexasAM, RBA, causal PINNs
%@article{wang2022and
{\bf Adaptive Weights:} The original PINNs used fixed weights in front of the various terms in the loss functions, requiring manual tuning.  
In \cite{wang2022and}, the authors used the Neural Tangent Kernel (NTK) for PINNs, which is a kernel that captures the behavior of neural networks in the infinite width limit during training via gradient descent. They proposed a novel gradient descent algorithm that utilizes the eigenvalues of the NTK to adaptively calibrate the loss weights and accelerate reduction of the total training error. 
%@article{mcclenny2020self
An even more effective method was proposed in \cite{mcclenny2020self} for multiscale systems and “stiff” PDEs. The loss weights are fully trainable
and applied to each training point individually, so the neural network learns which regions of the solution are stiff and focuses on them. The basic idea is to make the weights increase as the corresponding losses increase, which is accomplished by training the network to simultaneously
minimize the losses and maximize the weights (a min-max problem).
A variation of this idea that avoids computing the gradients but instead uses the residuals was proposed in \cite{anagnostopoulos2023residual}. 

% @article{anagnostopoulos2023residual
% A variation of this idea that avoids computing the gradients but instead uses the residuals was proposed in \cite{anagnostopoulos2023residual}.

%- CPINNs, XPINNs, Separable PINNs, also h-p PINNs

{\bf Domain Decomposition:} Another approach to tackle multiscale problems with PINNs is to combine them with domain decomposition methods as was done in \cite{jagtap2020conservative, jagtap2021extended}.In the first paper, domain decomposition was introduced for conservation laws, where a PINN is used for each
%%%%%%%%%%%%%%%%%%%%%%%%%%%%%%%%%%%%%%%%%%%%%%%%%%%%%
%%%%%%%%%%%%%%%%%%%%%%%%%%%%%%%%%%%%%%%%%%%%%%%%%%%%%
 subdomain and flux continuity at the interfaces together with state continuity (CPINN) are imposed in the loss function. In the second paper, the same idea was extended to arbitrary PDEs but the continuity of the residuals was enforced along the interfaces (XPINN). Both methods lead to effective scaling up of PINNs to large computational domains as well as parallel speed-up as each subdomain can be mapped to a separate GPU. However, XPINN in particular can be used for decomposing the temporal domain as well, hence leading to parallel-in-time computations of arbitrary PDEs. Another development inspired by domain decomposition and spectral elements in particular is the work on variational PINNs in \cite{kharazmi2021hp}. A general framework for hp-VPINNs was formulated  based on the nonlinear approximation of shallow and deep neural networks and hp-refinement via domain decomposition and projection onto the space of high-order polynomials. Specifically, the trial space is the space of neural network, which is defined globally over the entire computational domain, while the test space contains piecewise Legendre polynomials. 

%- Long-term integration: Kirby et al. sequential learning.
%@article{wang2022respecting
{\bf Long-Time Integration:} Another problematic issue in PINNs is the long-time integration of dynamical systems whose solution exhibits chaotic behavior. In 
\cite{wang2022and}, this shortcoming was attributed to the inability of existing PINNs formulations to respect the spatio-temporal causal structure that is inherent to the evolution of physical systems. They addressed it  by re-formulating the loss functions so that it can explicitly account for physical causality during model training. This approach still requires a very small temporal domain to be accurate, and various other attempts have resorted to sequential learning. A more general and unified approach was presented in \cite{penwarden2023unified}, where the authors  
 proposed a new stacked-decomposition method that bridges the gap between time-marching PINNs and XPINNs. They also introduced significant speed-ups by using transfer learning to initialize subnetworks in the domain and loss tolerance-based propagation for the subdomains. Moreover, they formulated a new time-sweeping collocation point algorithm inspired by the aforementioned PINNs causality. These methods form a unified framework, which overcomes training challenges in PINNs and XPINNs for time-dependent PDEs by respecting the causality in multiple forms and improving scalability by limiting the computation required per optimization iteration. 
 
%- PI-GANs for SPDES, fPINNs for fractional PINNs
% @article{pang2019fpinns,@article{yang2020physics,
{\bf Other Types of PDEs:} 
PINNs have also been applied to other types of PDES, e.g., in stochastic PDEs for uncertainty quantification, and in fractional PDEs for modeling anomalous transport. Specifically, 
in \cite{yang2020physics}, a new class of physics-informed generative adversarial networks (PI-GANs) was proposed to solve forward, inverse, and mixed stochastic problems in a unified manner based on a limited number of scattered measurements. Unlike standard GANs relying solely on data for training, they encoded into the architecture of Wassertein GANs the governing physical laws in the form of stochastic differential equations (SDEs) using automatic differentiation. The authors demonstrated the effectiveness of PI-GANs in solving SDEs for up to 120 dimensions, and suggested that PI-GANs could tackle very high dimensional problems given more sensor data with low-polynomial growth in computational cost.
To model anomalous transport in heterogeneous media, the authors of  \cite{pang2019fpinns} extended PINNs to fractional PINNs (fPINNs) to solve space-time fractional advection-diffusion equations, and studied  their convergence. A novel element of the fPINNs is the hybrid approach  introduced for constructing the residual in the loss function using both automatic differentiation for the integer-order operators and numerical discretization for the fractional operators. This approach bypasses the difficulties arising from the fact that automatic differentiation is not applicable to fractional operators because the standard chain rule in integer calculus is not valid in fractional calculus. Several examples were presented to identify the fractional orders, diffusion coefficients, and transport velocities, demonstrating accurate results even in the presence of significant noise.

%- theory: Molinaro and Shin
{\bf Theory:} Finally, we report on some early works on the theoretical foundations of PINNs. In \cite{shin2020convergence}, the authors demonstrated that, as the number of collocation points increases, the sequence of minimizers for PINNs – each corresponding to a sequence of neural networks – converges to the solution of the PDE for two classes of PDEs, linear second-order elliptic and parabolic. By adapting the Schauder approach and the maximum principle, they showed that the sequence of minimizers strongly converges to the PDE solution in $C^0$. Furthermore, they showed that if each minimizer satisfies the initial/boundary conditions, the convergence mode becomes $H^1$. 
In \cite{mishra2023estimates}, the authors provided upper bounds on the generalization error of PINNs approximating solutions of the forward problem for PDEs. They introduced an abstract formalism, and leveraged the stability properties of the underlying PDE to derive an estimate for the generalization error in terms of the training error and the number of training samples.

% \vspace{-0.7in} 
\section{Data-Driven Discovery of Dynamical Systems}
% \vspace{-0.5in} 

Next, we demonstrate a PINN-inspired approach for identifying nonlinear dynamical systems from data. Specifically, we blend classical tools from numerical analysis, namely the multi-step time-stepping schemes, with deep neural networks, to distill the mechanisms that govern the evolution of a given dataset. We test the effectiveness of our approach for a biomedical application example but more cases can be found in \cite{raissi2018multistep}.  In particular, we study the  glycolytic oscillator model as an example of complicated nonlinear dynamics typical of biological systems, and subsequently use symbolic regression to obtain the equations explicitly in analytical form.
% dynamical systems
%\section{Problem setup and solution methodology}\label{sec:setup}
Let us consider nonlinear dynamical systems of the form:
%\begin{equation}\label{eq:DynamicalSystems}
$\frac{d}{d t} \bm{x}(t) = \bm{f}\left(\bm{x}(t)\right)$,
%\end{equation}
where the vector $\bm{x}(t) \in \mathbb{R}^D$ denotes the state of the system at time $t$ and the function $\bm{f}$ describes the evolution of the system. Given noisy measurements of the state $\bm{x}(t)$ of the system at several time instances $t_1, t_2, \ldots, t_N$, our goal is to determine the function $\bm{f}$ and consequently discover the underlying dynamical system  from data. We proceed by applying the general form of a linear multistep method with  $M$ steps to equation %\eqref{eq:DynamicalSystems}
and obtain
%\begin{equation}\label{eq:multistep}
$\sum_{m=0}^M \left[\alpha_m \bm{x}_{n-m} + \Delta t \beta_m \bm{f}(\bm{x}_{n-m})\right] = 0, \ \ \ n = M, \ldots, N.$
%\end{equation}
Here, $\bm{x}_{n-m}$ denotes the state of the system $\bm{x}(t_{n-m})$ at time $t_{n-m}$. Different choices for the parameters $\alpha_m$ and $\beta_m$ result in specific schemes. 
%For instance, the trapezoidal rule 
%\begin{equation}\label{eq:trapezoidal}
%\bm{x}_n = \bm{x}_{n-1} + \frac{1}{2} \Delta{t} \left(\bm{f}(\bm{x}_n) + \bm{f}(\bm{x}_{n-1})\right),\ \ \ n = 1, \ldots, N,
%\end{equation}
%corresponds to the case where $M = 1$, $\alpha_0 = -1$, $\alpha_1 = 1$, and $\beta_0 = \beta_1 = 0.5$. 
We proceed by placing a neural network prior on the function $\bm{f}$. The parameters of this neural network can be learned by minimizing the mean squared error loss function
%\begin{equation}\label{eq:MSE}
$MSE := \frac{1}{N-M+1}\sum_{n=M}^{N} |\bm{y}_n|^2,$
%\end{equation}
where
%\begin{equation}\label{eq:multistep_NN}
$\bm{y}_n := \sum_{m=0}^M \left[\alpha_m \bm{x}_{n-m} + \Delta t \beta_m \bm{f}(\bm{x}_{n-m})\right], \ \ \ n = M, \ldots, N,$
%\end{equation}
is obtained from the multistep scheme.
%\section{Results}
%\subsection{Glycolytic oscillator}\label{sec:glycolytic}
As an example of complicated nonlinear dynamics typical of biological systems, we simulate the glycolytic oscillator model presented in \cite{daniels2015efficient}. The model consists of ordinary differential equations for the concentrations of 7 biochemical species, see \cite{raissi2018multistep}. The parameters of the model are chosen according to table 1 of \cite{daniels2015efficient}.

\allowdisplaybreaks
\begin{align}
\small
\frac{dS_1}{dt} &= J_0 - \frac{k_1 S_1 S_6}{1 + (S_6/K_1)^q}, \nonumber \\
\frac{dS_2}{dt} &= 2\frac{k_1 S_1 S_6}{1 + (S_6/K_1)^q} - k_2 S_2 (N - S_5) - k_6 S_2 S_5, \nonumber \\
\frac{dS_3}{dt} &= k_2 S_2 (N - S_5) - k_3 S_3 (N - S_6), \nonumber \\
\frac{dS_4}{dt} &= k_3 S_3 (A - S_6) - k_4 S_4 S_5 - \kappa (S_4 - S_7), \label{eq:glycolytic} \\
\frac{dS_5}{dt} &= k_2 S_2 (N - S_5) - k_4 S_4 S_5 - k_6 S_2 S_5, \nonumber \\
\frac{dS_6}{dt} &= -2\frac{k_1 S_1 S_6}{1 + (S_6/K_1)^q} + 2 k_3 S_3 (A - S_6) - k_5 S_6, \nonumber \\
\frac{dS_7}{dt} &= \psi \kappa (S_4 - S_7) - k S_7. \nonumber
\end{align}
 %As shown in figure \ref{fig:Glycolytic}, 
Data from the simulation are collected from $t = 0$ to $t = 10$ with a time-step size of $\Delta t = 0.01$. We employ a DNN with one hidden layer and $256$ neurons to represent the nonlinear dynamics. As for the multi-step scheme, we use Adams-Moulton with $M=1$ steps. Upon training the DNN, we solve the identified system using the same initial condition as the ones used for the exact system. 
%As depicted in figure \ref{fig:Glycolytic}, 
As shown in \cite{raissi2018multistep}, the learned system correctly captures the form of the dynamics. Here, we use symbolic regression, a method that merges genetic programming with machine learning, to identify mathematical expressions that closely align with our dataset. Initially, this method involves creating a varied pool of potential equations, each expressed mathematically using basic operations (e.g., addition, subtraction, multiplication, and division) and various functions (like addition, subtraction, multiplication, and division). The adequacy of each equation is assessed against the data using a specific fitness function, generally based on the mean squared error (MSE). Through the use of genetic algorithms, the most effective equations are carried forward, undergoing genetic processes such as crossover (mixing elements of two equations) and mutation (altering parts of an equation) to generate new candidates. This cycle of generation and refinement proceeds until certain criteria are met, like reaching a pre-defined number of generations or achieving a particular level of fitness. The results, including comparisons of these equations with the exact right-hand side equations of the ODEs as seen in Eq. \ref{eq:glycolytic}, are displayed in Table \ref{table:SR}. The notably low relative errors indicate that the system learned through Symbolic Regression successfully mirrors the system's inherent dynamics. We opted for PySR package proposed by  \cite{cranmer2023interpretable} over gplearn as PySR has proven to be a more robust and efficient framework, as discussed in \cite{ahmadi2023aiaristotle}.\\

\begin{table}[h!]
\centering
\resizebox{\textwidth}{!}{% Resize table to fit within the text width
\begin{tabular}{lc|ccc}
\hline
\textbf{Eq.}  & \textbf{PySR}   & \textbf{True Expression} & \textbf{RE} \\
\hline \hline
1\textsuperscript{st} ODE  &  $2.6 - \frac{100 S_{1} S_{6}}{38.3 S_{6}^{3} - 33.7 S_{6}^2 + 10.5S_{6}}$    &  $2.5 - \frac{100 S_1 S_6}{1 + \left(\frac{S_6}{0.52}\right)^4}$ & $8.04e-02$ \\
7\textsuperscript{th} ODE  &  $1.3S_4 - 3.1S_7$    &  $1.3S_4 - 3.1S_7$ & $0.00$ \\
Part of 5\textsuperscript{th} ODE  &  $5.99S_2 - 18.0S_2S_5$    &  $6.0S_2 - 18.0S_2 S_5$ & $3.38e-03$ \\
\hline
\end{tabular}}
\caption{\small Glycolytic oscillator model: Results of symbolic regression for symbolic expression discovery using the PySR package. `RE' represents the Relative Error computed between the exact function and the function inferred by PySR.}
\label{table:SR}
\end{table}

%\begin{figure}[!t]
%\includegraphics[width = 1.0\textwidth]{Glycolytic.pdf}
%\caption{{\em Glycolytic oscillator:} Exact versus learned dynamics for %random initial conditions chosen from the ranges provided in table 2 of %\cite{daniels2015efficient}.}
%\label{fig:Glycolytic}
%\end{figure}

\vspace{-0.35in} 
\section{Outlook}\label{sec:Conclusion} 
PINNs have been used so far all across the scientific and engineering fields,
from geophysics and astrophysics to engineering design, digital twins, computational mechanics, biomedical engineering and even in quantitative pharmacology as shown in the example above. Compared to finite elements that it took several decades to be adopted as a mainstream computational tool, PINNs have already been adopted by the industry as they remove the ``tyranny" of mesh generation, they blend seamlessly data and physics, and can even discover new governing equations from data as in the aforementioned example. However,  there are still many open issues to be resolved, including the low-accuracy compared to high-order numerical methods, the computational cost, which is excessive especially for forward problems, and scalability to high dimensions. All these problems are currently being investigated by many research groups around the world and there has been great progress on all fronts already, e.g., using tensor type DNNs accelerates PINNs by two orders of magnitude.

\vspace{-0.2in} 
\address{$^{1}$University of California, Riverside, Department of Mathematics, Riverside, CA 92521, USA \email{maziar.raissi@gmail.com}}

\address{$^{2}$Department of Mechanical Engineering and Applied Mechanics, University of Pennsylvania, Philadelphia, PA 19104, USA \email{pgp@seas.upenn.edu}}

\address{$^{3}$Center for Biomedical Engineering, School of Engineering, Brown University, Providence, RI 02912, USA \email{nazanin@brown.edu}}

\address{$^{4}$Division of Applied Mathematics and School of Engineering, Brown University, Providence, RI 02912, USA \\ \email{george\_karniadakis@brown.edu}}


\begin{thebibliography}{00}
\bibitem[Raissi et al. (2017a)]{raissi2017physics_I} M. Raissi, P. Perdikaris, G. E. Karniadakis,
\emph{Physics Informed Deep Learning (Part II): Data-driven Discovery of Nonlinear Partial Differential Equations}, 
arXiv preprint arXiv:1711.10566 (2017).

\bibitem[Raissi et al. (2017b)]{raissi2017physics_II} M. Raissi, P. Perdikaris, G. E. Karniadakis,
\emph{Physics Informed Deep Learning (Part I): Data-driven Solutions of Nonlinear Partial Differential Equations}, 
arXiv preprint arXiv:1711.10561 (2017).

\bibitem[Raissi et al. (2019)]{raissiJCP2019} M. Raissi, P. Perdikaris, G. E. Karniadakis,
\emph{Physics-informed neural networks: A deep learning framework for solving forward and inverse problems involving nonlinear partial differential equations}, 
Journal of Computational Physics {\bf 378}, (2019), 686--707.

\bibitem[Wang et al. (2022)]{wang2022and} S. Wang, X. Yu, P. Perdikaris,
\emph{When and why PINNs fail to train: A neural tangent kernel perspective}, 
Journal of Computational Physics, {\bf 449} (2022), 110768.

\bibitem[McClenny and Braga-Neto. (2020)]{mcclenny2020self} L. McClenny, U. Braga-Neto,
\emph{Self-adaptive physics-informed neural networks using a soft attention mechanism}, 
arXiv preprint arXiv:2009.04544 
 (2020).

\bibitem[Anag. et al. (2023)]{anagnostopoulos2023residual} S. Anagnostopoulos, J. D. Toscano, N. Stergiopulos, G. E. Karniadakis
\emph{Residual-based attention and connection to information bottleneck theory in PINNs}, 
arXiv preprint arXiv:2307.00379 (2023).

\bibitem[Jagtap et al. (2020)]{jagtap2020conservative} A. D. Jagtap, E. Kharazmi, G. E. Karniadakis,
\emph{Conservative physics-informed neural networks on discrete domains for conservation laws: Applications to forward and inverse problems}, 
Computer Methods in Applied Mechanics and Engineering, {\bf 365} (2020), 113028.

\bibitem[Jagtap et al. (2021)]{jagtap2021extended} A. D. Jagtap, G. E. Karniadakis,
\emph{Extended Physics-informed Neural Networks (XPINNs): A Generalized Space-Time Domain Decomposition based Deep Learning Framework for Nonlinear Partial Differential Equations.}, 
AAAI spring symposium: MLPS, {\bf 10} (2021).

\bibitem[Kharazmi et al. (2021)]{kharazmi2021hp} E. Kharazmi, Z. Zhang, G. E. Karniadakis,
\emph{hp-VPINNs: Variational physics-informed neural networks with domain decomposition}, 
Computer Methods in Applied Mechanics and Engineering, {\bf 374}, 113547, (2021).


\bibitem[Penwarden et al. (2023)]{penwarden2023unified} M. Penwarden, A. D. Jagtap, S. Zhe, G. E. Karniadakis, R.M. Kirby,
\emph{A unified scalable framework for causal sweeping strategies for Physics-Informed Neural Networks (PINNs) and their temporal decompositions}, 
arXiv preprint arXiv:2302.14227, (2023).


\bibitem[Yang et al. (2020)]{yang2020physics} L. Yang, D. Zhang, G. E. Karniadakis,
\emph{Physics-informed generative adversarial networks for stochastic differential equations}, 
SIAM Journal on Scientific Computing, {\bf 42 (1)}, A292--A317, (2020).

\bibitem[Pang et al. (2019)]{pang2019fpinns} G. Pang, L. Lu, G. E. Karniadakis,
\emph{fPINNs: Fractional physics-informed neural networks}, 
SIAM Journal on Scientific Computing, {\bf 41 (4)}, A2603--A2626, (2019).

\bibitem[Shin et al. (2019)]{shin2020convergence} Y. Shin, J. Darbon, G. E. Karniadakis,
\emph{On the convergence of physics informed neural networks for linear second-order elliptic and parabolic type PDEs}, 
arXiv preprint arXiv:2004.01806, (2020).

\bibitem[Mishra and Molinaro (2019)]{mishra2023estimates} S. Mishra, R. Molinaro,
\emph{Estimates on the generalization error of physics-informed neural networks for approximating PDEs}, 
IMA Journal of Numerical Analysis, {\bf 43 (1)}, 1--43, (2023).


\bibitem[Raissi et al. (2018)]{raissi2018multistep} M. Raissi, P. Perdikaris, G. E. Karniadakis,
\emph{Multistep neural networks for data-driven discovery of nonlinear dynamical systems}, 
arXiv preprint arXiv:1801.01236, (2018).


\bibitem[Daniels and Ilya. (2015)]{daniels2015efficient} B. C. Daniels, I. Nemenman,
\emph{Efficient inference of parsimonious phenomenological models of cellular dynamics using S-systems and alternating regression}, 
PloS one, {\bf 10 (3)}, e0119821, (2015).


%\bibitem[Brunton et al., (2016)]{brunton2016discovering} S. L. Brunton, J. L. Proctor, J. N. Kutz,
%\emph{Discovering governing equations from data by sparse identification of nonlinear dynamical systems}, 
%Proceedings of the National Academy of Sciences, {\bf 113 (15)}, 3932--3937, (2016).


%\bibitem[Goodfellow et al., (2016)]{goodfellow2016deep} I. Goodfellow, Y. Bengio, A. Courville,
%\emph{Deep learning}, 
%MIT press, {\bf 113 (15)}, 3932--3937, (2016).

%\bibitem[Chang et al. (2017)]{chang2017multi} B. Chang, L. Meng, E. Haber, F. Tung, D. Begret
%\emph{Multi-level Residual Networks from Dynamical Systems View}, 
%arXiv preprint arXiv:1710.10348, (2017).

%\bibitem[Lu et al. (2017)]{lu2017beyond} Y. Lu, A. Zhong, Q. Li, D. Bin
%\emph{Beyond Finite Layer Neural Networks: Bridging Deep Architectures and Numerical Differential Equations}, 
%arXiv preprint arXiv:1710.10121, (2017).

%\bibitem[Raissi et al., (2020)]{raissi2020hidden} M. Raissi, A. Yazdani, G. E. Karniadakis,
%\emph{Hidden fluid mechanics: Learning velocity and pressure fields from flow visualizations}, 
%Science, {\bf 367 (6481)}, 1026--1030, (2020).
\bibitem[Ahmadi Daryakenari et al. (2023)]{ahmadi2023aiaristotle} N. Ahmadi Daryakenari, M. De Florio, K. Shukla, G. E. Karniadakis
\emph{AI-Aristotle: A Physics-Informed framework for Systems Biology Gray-Box Identification},
arXiv preprint arXiv:2310.01433 (2023).

\bibitem[Cranmer (2023)]{cranmer2023interpretable} M. Cranmer,
\emph{Interpretable machine learning for science with PySR and SymbolicRegression. jl},
arXiv preprint arXiv:2305.01582, (2023).

\end{thebibliography}
\end{document}